\documentclass[10pt,twocolumn,letterpaper]{article}

\usepackage[pagenumbers]{cvpr} % To force page numbers, e.g. for an arXiv version

\usepackage{graphicx}
\usepackage{amsmath}
\usepackage{amssymb}
\usepackage{booktabs}
\usepackage{multirow, adjustbox}
\usepackage{xcolor,colortbl}
\usepackage{microtype}
\usepackage{enumitem}

\usepackage[pagebackref,breaklinks,colorlinks]{hyperref}
\usepackage{etoolbox}
\usepackage{pifont}
\usepackage[capitalize]{cleveref}
\crefname{figure}{Fig.}{Figs.}
\crefname{figure}{Figure}{Figures}
\crefname{section}{Sec.}{Secs.}
\Crefname{section}{Section}{Sections}
\Crefname{table}{Table}{Tables}
\crefname{table}{Tab.}{Tabs.}

\DeclareMathOperator*{\argmax}{arg\,max}

\def\cvprPaperID{9117} % *** Enter the CVPR Paper ID here
\def\confName{CVPR}
\def\confYear{2023}

\begin{document}
\setlength{\tabcolsep}{3pt}
\setlength{\abovedisplayskip}{6pt}
\setlength{\belowdisplayskip}{6pt}
\definecolor{Gray}{gray}{0.85}
\newcolumntype{g}{>{\columncolor{Gray}}c}
\newcommand{\cmark}{\ding{51}}
\newcommand{\xmark}{\ding{55}}
\title{Guiding Pseudo-labels with Uncertainty Estimation for Source-free Unsupervised Domain Adaptation}

\author{Mattia Litrico \hspace{1cm} Alessio Del Bue \hspace{1cm} Pietro Morerio\\
Pattern Analysis and Computer Vision (PAVIS) - Istituto Italiano di Tecnologia\\
{\tt\small mattia.litrico@phd.unict.it, \tt\small alessio.delbue@iit.it, \tt\small pietro.morerio@iit.it}
}
\maketitle

%%%%%%%%% ABSTRACT
\begin{abstract}
Standard Unsupervised Domain Adaptation (UDA) methods assume the availability of both source and target data during the adaptation. In this work, we investigate the Source-free Unsupervised Domain Adaptation (SF-UDA), a specific case of UDA where a model is adapted to a target domain without access to source data. We propose a novel approach for the SF-UDA setting based on a loss reweighting strategy that brings robustness against the noise that inevitably affects the pseudo-labels.
The classification loss is reweighted based on the reliability of the pseudo-labels that is measured by estimating their uncertainty. Guided by such reweighting strategy, the pseudo-labels are progressively refined by aggregating knowledge from neighbouring samples. 
Furthermore, a self-supervised contrastive framework is leveraged as a target space regulariser to enhance such knowledge aggregation. A novel negative pairs exclusion strategy is proposed to identify and exclude negative pairs made of samples sharing the same class, even in presence of some noise in the pseudo-labels.
Our method outperforms previous methods on three major benchmarks by a large margin. We set the new SF-UDA state-of-the-art on VisDA-C and DomainNet with a performance gain of +1.8\% on both benchmarks and on PACS with +12.3\% in the single-source setting and +6.6\% in\ multi-target adaptation. Additional analyses demonstrate that the proposed approach is robust to the noise, which results in significantly more accurate pseudo-labels compared to state-of-the-art approaches.
\end{abstract}

%%%%%%%%% BODY TEXT
\section{Introduction}
\label{sec:intro}
Deep learning methods achieve remarkable performance in visual tasks when the training and test sets share a similar distribution, while their generalisation ability on unseen data decreases in presence of the so called  \textit{domain shift} \cite{domain_shift, domain_shift2}. Moreover, DNNs require a huge amount of labelled data to be trained on a new domain entailing a considerable cost for collecting and labelling the data.
Unsupervised Domain Adaptation (UDA) approaches aim to transfer the knowledge learned on a labelled source domain to an unseen target domain without requiring any target label \cite{uda1,uda2,uda3,uda4}.     

Common UDA techniques have the drawback of requiring access to source data while they are adapting the model to the target domain. This may not always be possible in many applications, i.e. when data privacy or transmission bandwidth become critical issues. In this work, we focus on the setting of Source-free Unsupervised Domain Adaptation (SF-UDA) \cite{sfda,sfda1,sfda2,sfda3}, also referred to as source-free UDA, where \textbf{source data is no longer accessible} during adaptation, but only unlabelled target data is available. As a result, standard UDA methods cannot be applied in the SF-UDA setting, since they require data from both domains.

Recently, several SF-UDA methods have been proposed, focusing on generative models \cite{generative,generative2}, class prototypes \cite{sfda2}, entropy-minimisation \cite{sfda2,sfda1}, self-training \cite{sfda2} and auxiliary self-supervised tasks \cite{sfda}. Yet, generative models require a large computational effort to generate images/features in the target style \cite{generative}, entropy-minimisation methods often lead to posterior collapse \cite{sfda2}  and the performance of self-training solutions \cite{sfda2} suffer from noisy pseudo-labels. Self supervision and pseudo-labelling have also been introduced as joint SF-UDA strategies \cite{sfda, ontarget}, raising the issue of choosing a suitable pretext task and refining pseudo-labels. For instance, Chen et al. \cite{contrastive_testtime} propose to refine predictions within a self-supervised strategy. Yet, their work does not take into account the noise inherent in pseudo-labels, which leads to equally weighting all samples without explicitly accounting for their uncertainty. This may cause pseudo-labels with high uncertainty to still contributing in the classification loss, resulting in detrimental noise overfitting and thus poor adaptation.

\begin{figure*}[ht!]
\centering
\includegraphics[width=1.8\columnwidth]{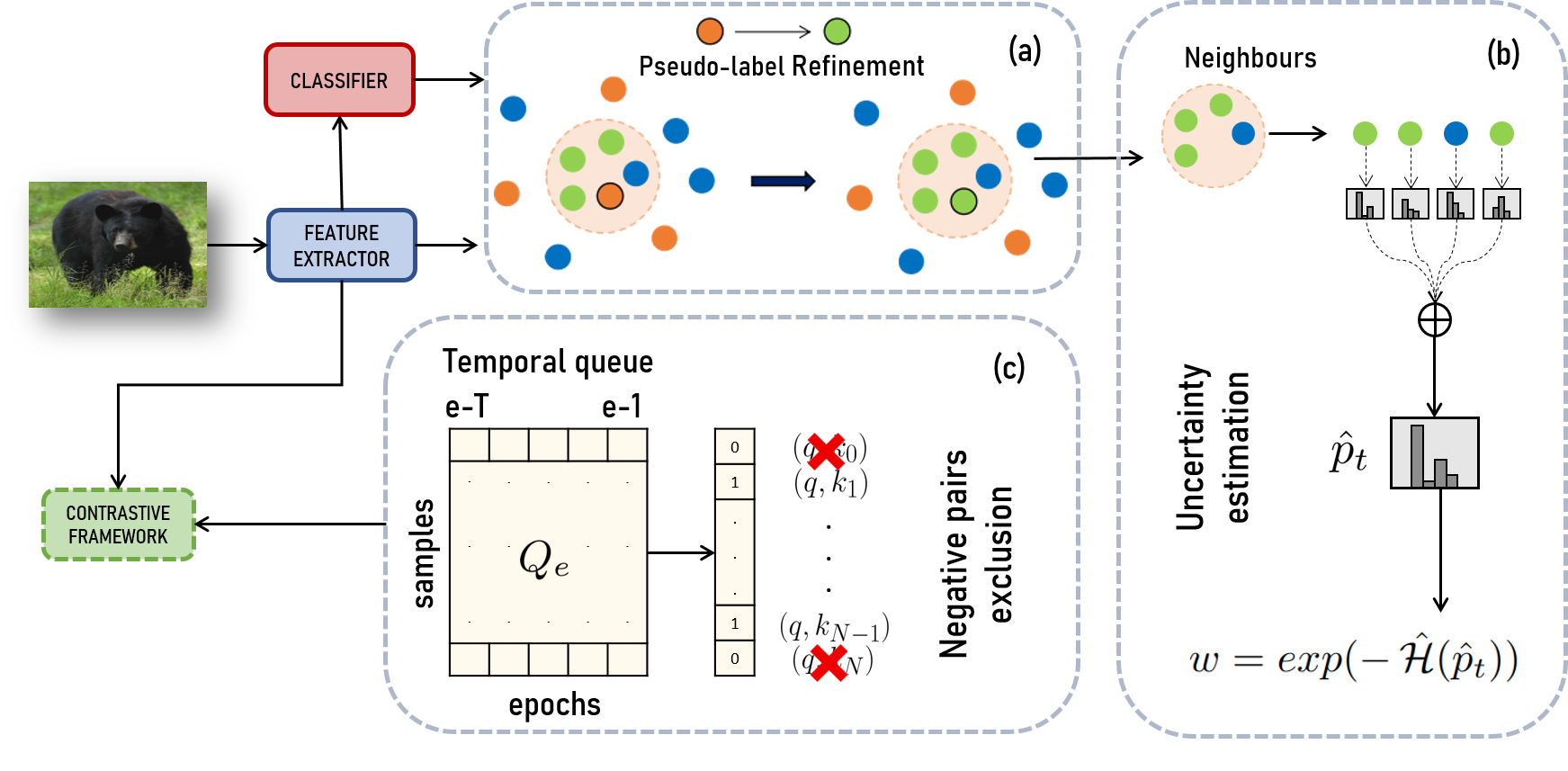}
\caption{(a) We obtain the refined pseudo-label $\hat{y}_t$ (green circle with black outline) for the current sample by looking at pseudo-labels of neighbour samples. (b) Predictions from neighbours are used to estimate the uncertainty of $\hat{y}_t$ by computing the weight $w$ through Entropy $\hat{\mathcal{H}}$. (c) A temporal queue $Q_e$ storing predictions at past T epochs, i.e. $\{e-T,...,e-1 \}$, is used within the contrastive framework to exclude pairs of samples sharing the same class from the list of negative pairs \textit{(query, key)}.}
\label{fig:main_figure}
\vspace{-1.0em}
\end{figure*}

In this work, we propose a novel Source-free adaptation approach that builds upon an initial pseudo-labels assignment (for all target samples) performed by using the pretrained source model, always assumed to be available. 
To obtain robustness against the noise that inevitably affects such pseudo-labels, we propose to reweight the classification loss based on the reliability of the pseudo-labels. We measure it by estimating pseudo-labels uncertainty, after they have been refined by knowledge aggregation from neighbours sample, as shown in \cref{fig:main_figure} (a).
The introduced loss reweighting strategy penalises pseudo-labels with high uncertainty to guide the learning through reliable pseudo-labels.
Differently from previous reweighting strategies, we reweight the loss by estimating the uncertainty of the refined pseudo-labels by simply analysing neighbours' predictions, as shown in \cref{fig:main_figure} (b).

The process of refining pseudo-labels necessarily requires a regularised target feature space in which neighbourhoods are composed of semantically similar samples, possibly sharing the same class. With this objective, we exploit an auxiliary self-supervised contrastive framework. 

Unlike prior works, we introduce a novel negative pairs exclusion strategy that is robust to noisy pseudo-labels, by leveraging past predictions stored in a temporal queue. This allows us to identify and exclude negative pairs made of samples belonging to the same class, even if their pseudo-labels are noisy, as shown in \cref{fig:main_figure} (c).

We benchmark our method on three major domain adaptation datasets outperforming the state-of-the-art by a large margin. Specifically, on VisDA-C and DomainNet, we set the new state-of-the-art with 90.0\% and 69.6\% accuracy, with a gain of +1.8\% in both cases, while on PACS we improve the only existing SF-UDA baseline \cite{nel} by +12.3\% and +6.6\% in single-source and multi-target settings. Ablation studies demonstrate the effectiveness of individual components of our pipeline in adapting the model from the source to the target domain. We also show how our method is able to progressively reduce the noise in the pseudo-labels, better than the state-of-the-art. 

To summarise, the main contributions of this work are:
\begin{itemize}[noitemsep,nolistsep]
  \item We introduce a novel loss re-weighting strategy that evaluates the reliability of refined pseudo-labels by estimating their uncertainty. This enables our method to mitigate the impact of the noise in the pseudo-labels. To the best of our knowledge, this is the first work that estimates the reliability of pseudo-labels after their refinement.
  \item We propose a novel negative pairs exclusion strategy which is robust to noisy pseudo-labels, being able to identify and exclude those negative pairs composed of same-class samples.
  
  \item We validate our method on three benchmark datasets, outperforming SOTA by a large margin, while also proving the potential of the approach in progressively refining the pseudo-labels. 
\end{itemize}

The remainder of the paper is organized as follows. In Section~\ref{sec:relatedwork}, we discuss related works in the literature. Section~\ref{sec:method} describes the proposed method. Section~\ref{sec:experiments_results} illustrates the experimental setup and reports the obtained results. Finally, limitations are discussed in Section \ref{sec:limit} and conclusions are drawn in Section~\ref{sec:conclusion}.

\section{Related Work}
\label{sec:relatedwork}
    
\textbf{Domain Adaptation.}
    Standard Unsupervised Domain Adaptation (UDA) methods aim to adapt a model trained on a source domain, in order to work also on an unseen target domain, when domain shift is present between the two. Many early methods relied on aligning statistics of the distributions \cite{domainnet}, reducing the discrepancy between domains \cite{mmd1,mmd2,adversarial,uda2,uda3, uda4}, as well as exploiting generative models \cite{uda3}. However, all these methods required the access to both source and target data during the adaptation.
    Recently, some Source-free adaptation methods have been proposed, adapting to the target domain using only the source model and unlabelled target data \cite{sfda,sfda1,sfda2,sfda3,generative,generative2,ontarget,contrastive_testtime}. TENT \cite{sfda1} and SHOT \cite{sfda2} introduced entropy minimisation and pseudo-labeling. On-target \cite{ontarget} proposed to combine a pseudo-labels generation and a self-supervised task, but the pseudo-labels are not refined during the adaptation. Instead, \cite{contrastive_testtime} proposed a self-supervised strategy to refine the pseudo-labels. The main limitation is that all the pseudo-labels equally contributed to the loss, without considering the noise that inevitably affects the (possibly refined) pseudo-labels. Moreover, it did not employ any other countermeasure to mitigate the effects of this noise.\vspace{3pt}
%%%%%%%%%%%
\textbf{\\Self-supervised Learning and Pseudo Labeling.}
    Self-supervised methods are successful in learning transferable representations of visual data \cite{self_supervised1, cl1, self_supervised2, cl2, self_supervised3, cl3, moco, self_supervised4, self_supervised5}. Specifically, \cite{cl1,cl2,cl3} showed how contrastive-based pretext tasks could help in enhancing the generalisation ability of deep models. Moreover, some self-supervised approaches have been recently exploited in both UDA \cite{self_supervised6, self_supervised7} and SF-UDA \cite{sfda,ontarget,contrastive_testtime} settings.
    Pseudo-labeling is a simple but effective technique used in semi-supervised learning \cite{pseudolabel_1, fixmatch}, self-supervised learning \cite{self_supervised1} and domain adaptation \cite{sfda2, pseudolabel_2, ontarget, contrastive_testtime}. It consists in using labels predicted by the model as self-supervision. Fix-Match \cite{fixmatch} and On-target \cite{ontarget} are methods that used pseudo-labels but they did not perform any labels refinement. In this work, we use both self-supervision and pseudo-labeling as an approach to exploit the structure of the target features space to progressively refine pseudo-labels.\vspace{3pt}
%%%%%%%%%%%
\textbf{\\Learning with Noisy Labels.}
%TODO discutere dividemix
    While deep learning models achieve competitive results with carefully labelled data, their performance decreases when they are trained with noisy labels. Zhang et al. \cite{noise_overfitting} demonstrated that a deep neural network can easily overfit an entire dataset with any ratio of corrupted labels which results in poor generalisation on test data. To address this problem, different approaches have been proposed focusing on the creation of noise-robust losses \cite{noise_robust_loss, MAEGhosh, NCE, NLNL}, the estimation of the noise-transition matrix \cite{noise_transition_matrix}, the selection of clean from noisy samples \cite{sample_selection, DivideMix}, as well as the reweighting of the loss based on the reliability of the given label \cite{reweighting, reweighting3, reweighting4}. Works in \cite{reweighting2} and \cite{reweighting3} proposed to reweight the loss based on weights learned using a meta-learned curriculum, but they require a noisy-free validation set which does not fit the SF-UDA setting. Furthermore, they do not refine noisy labels meaning that the amount of noise in the labels will not be reduced during training. Instead, we propose a reweighting strategy that estimates the uncertainty (reliability) of refined pseudo-labels based on the consensus among neighbours. In addition, our reweighting strategy does not require either noise-free validation set or target labels. Last, NEL \cite{nel} combined a \textit{Negative Learning} loss with a pseudo-labels refinement framework based on ensembling. Negative Learning  \cite{NLNL} refers to an indirect learning method which uses complementary labels to combat noise. While we also exploit a Negative Learning loss, we do not require an ensemble of networks in order to perform refinement, which results in a large computational cost.

\section{Proposed Method}
\label{sec:method}

    This work addresses the Source-free adaptation problem for the task of image classification.
    Let $\mathcal{D}_s$ be the source data composed by pairs $\{x_s, y_s\} $, where $x_s \in \mathcal{X}_s$ and $y_s \in \mathcal{Y}_s$ are images and ground truth labels, respectively.
    Let $\mathcal{D}_t$ be the target data composed by only images $\{x_t^i\}_{i=1}^N$, where  $x_t^i \in \mathcal{X}_t$. The underlying labels $y_t^i \in \mathcal{Y}_t$ are available only for evaluation purposes. In the SF-UDA setting, the source data $\mathcal{D}_s$ cannot be used during adaptation. Regardless of this limitation, given the trained source model only, we adapt the model to work on the unlabelled target data $\mathcal{D}_t$. 
    
    The model has a typical architecture composed by a features extractor $f_s: \mathcal{X}_s \rightarrow \mathbb{R}^P$ and a classifier $h_s: \mathbb{R}^P \rightarrow \mathbb{R}^C$, where $P$ is the length of features vectors and $C$ is the number of classes.
    
    At the beginning of the adaptation process, the source model is used to generate pseudo-labels for each of the unlabelled target image $x_t^i$. Due to the domain shift between source and target domains, the source model makes a consistent amount of incorrect assignments, which can be interpreted as noise in pseudo-labels. The goal of the adaptation phase is thus to progressively refine the noisy pseudo-labels, which in turn results to progressively adapt the source model to the target domain.

    \subsection{Pseudo-label Refinement via Nearest neighbours Knowledge Aggregation}
    \label{subsubsec:nnvoting}
    
    Similar to \cite{contrastive_testtime}, the refinement of the pseudo-labels is accomplished by aggregating knowledge from nearest neighbours samples. The underlying idea is that similar samples are likely to have the same label. Moreover, here we assume that features from semantically similar images should lie close in the feature space. This assumption is satisfied by employing a contrastive framework that pulls close features from similar samples. The strategy used to aggregate the knowledge from neighbours is pictorially shown in \cref{fig:main_figure} (a) and it is based on soft voting.
    
    More formally, given a target image $x_t$ and a weak augmentation $t_{wa}$ drawn randomly from the distribution $\mathcal{T}_{wa}$, we obtain a features vector $z = f_t(t_{wa}(x_t))$ from the weakly augmented image $t_{wa}(x_t)$. The features $z$ are then used to search the neighbours of the sample $x_t$ in the target features space. Consequently, the pseudo-label of $x_t$ is refined by aggregating knowledge from the selected neighbours. 
    To this aim, the probability outputs from the selected neighbours are averaged to perform a soft-voting strategy \cite{softvoting}:
    \begin{equation}
        \hat{p}_t^{(c)} = \frac{1}{K}\sum_{i \in \mathcal{I}} p'^{(c)}_i,
    \end{equation}
    where $\mathcal{I}$ is the set of indices of the selected neighbours and the superscript $c$ indicates that the averaging operation is performed for each class.
    To obtain a refined pseudo-label, we use the argmax operation upon $\hat{p}_t$:
    \begin{equation}
        \hat{y_t} = \argmax_c \; \hat{p}_t^{(c)}.
    \end{equation}
    The refined pseudo-labels are then used as self-supervision for target samples (see \cref{subsec:overallframework}).
    
    The aforementioned refining process needs a representation of the target features space where to search for neighbour samples. This is allowed by a bank $B$ of length $M$, which stores pairs $\{z'_j, p_j'\}_{j=1}^M$ of features and predictions obtained from weakly augmented target samples that are selected randomly from the target dataset. The neighbours are then selected by computing the cosine distances between the features of $x_t$ and features stored in the bank.
    The $K$ samples with the lowest distance are selected as neighbours.
    Following \cite{moco}, to maintain the information stored in the bank more stable, we use a slowly changing momentum model $g'_t (\cdot) = h'_t(f'_t(\cdot))$ to update features $z'$ and predictions $p'$. 
    
    \begin{figure}[t!]
    \includegraphics[width=\columnwidth]{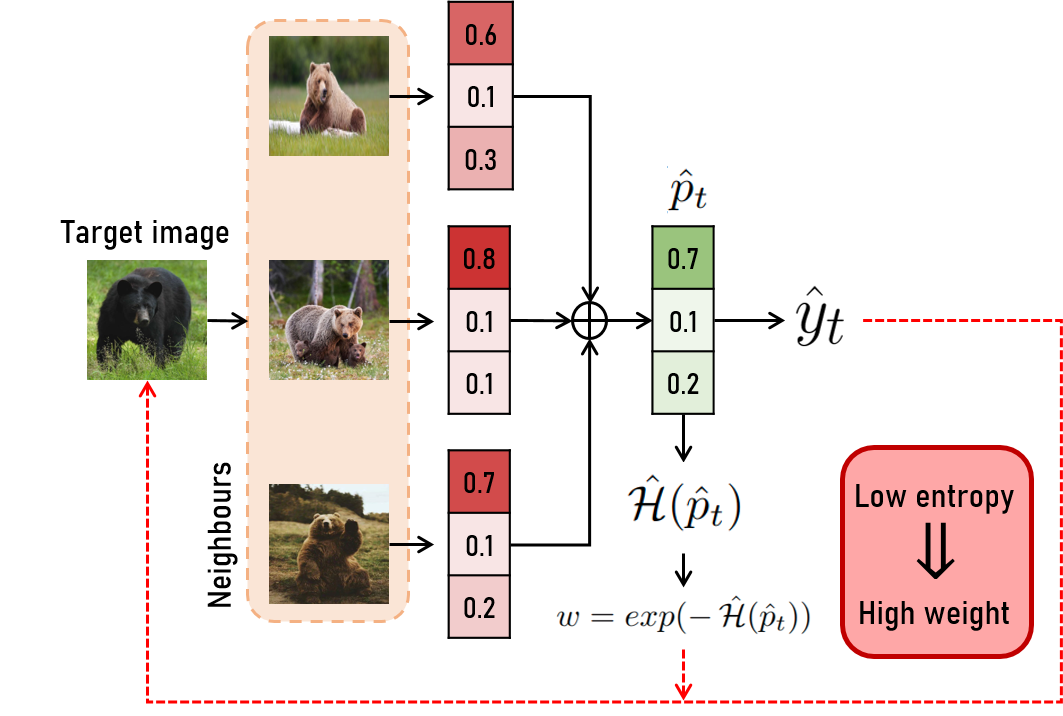}
    \centering
        \vspace{-10pt}
    \caption{We average prediction scores of the neighbour samples (red vectors) to obtain the average score vector $\hat{p}_t$. If the network consistently predicts the same class for neighbour samples, $\hat{p}_t$ has low entropy and we thus assign a high weight to the refined pseudo-label $\hat{y}_t$, considering it reliable.}
    \label{fig:uncertainty_estimation}
    \vspace{-10pt}
    \end{figure}

    \subsection{Loss Reweighting with Pseudo-labels Uncertainty Estimation}
    \label{subsec:entropy_estimation}
    The refined pseudo-labels $\hat{y}_t$ are used as a self-supervised signal for computing a standard classification loss on target data, as explained in \cref{subsec:overallframework}.
    However, since the refinement is an iterative process, the refined pseudo-labels obtained with neighbours' knowledge aggregation still contain some noise. Therefore, equally weighting all the pseudo-labels will disrupt the adaptation, since the model is trained with incorrect labels. 
    To solve this issue, we propose a novel way to reweight the classification loss by estimating the uncertainty of pseudo-labels after their refinement. Such estimation is performed by considering the neighbours only, disregarding the pseudo-label of the sample itself. We built upon the intuition that since pseudo-labels are obtained from neighbours' predictions, their uncertainty can be determined by the neighbours' accuracy. But since we do not have any target label, we cannot compute the neighbours' accuracy. In this section, we thus want to answer the following question: \textit{Can we empirically estimate the uncertainty of a pseudo-label using only neighbours' predictions?}  
    
    To answer this question, we introduce a method based on entropy-based uncertainty estimation using the consensus among neighbours' predictions. The underlying idea is that if the network predicts the same class for the neighbour samples, we could consider the derived pseudo-labels reliable (low uncertainty). Otherwise, if neighbours' predictions mostly disagree with each other, the obtained pseudo-labels should be considered unreliable (high uncertainty).
    Moreover, we observe that the averaged scores vector $\hat{p}_t$, obtained by averaging neighbours' probabilistic outputs, has low entropy in the former case and high entropy in the latter case, as illustrated in the example of \cref{fig:uncertainty_estimation}. Hence, we reweight the classification loss in \cref{eq:cls_loss} computing a weight $w$ that puts more importance on pseudo-labels obtained from $\hat{p}_t$ with low entropy and less importance to pseudo-labels obtained from $\hat{p}_t$ with high entropy.
    
    More formally, given a target sample $x_t$ we obtain the averaged scores vector $\hat{p}_t$ from the soft-voting strategy, as explained in \cref{subsubsec:nnvoting}. Note that $\hat{p}_t$ is obtained by averaging probability distributions, so it is still a probability distribution. Then, we compute the entropy of $\hat{p}_t$ as: 
    \begin{equation}
    \mathcal{H}(\hat{p}_t) = \mathbb{E}[I(\hat{p}_t)] = - \, \sum_{c=1}^C \hat{p}_t^c \, \log_2 \, \hat{p}_t^c,
    \end{equation}
    where $C$ is the number of classes in the dataset.
    Entropy is also re-scaled by its maximum as follows:
    \begin{equation}
        \hat{\mathcal{H}}(\hat{p}_t) = \frac{\mathcal{H}(\hat{p}_t)}{\log_2 \, C}.
    \end{equation}
    % in order to account for  its maximum value ($log_2 \, C$) .
    From the normalised entropy value $\hat{\mathcal{H}}(\hat{p}_t)$, we obtain the weight $w$ for the sample $x_t$ as:
    \begin{equation}
    \label{eq:reweighting}
    w_{x_t} = exp(- \, \hat{\mathcal{H}}(\hat{p}_t)).
    \end{equation}
    
    The motivation behind the negative exponential function is twofold. First, the negative sign is used to invert the behaviour of the exponential function to put more weight on low entropy values and less weight on high entropy values. Then, the exponential function does not penalise too much samples near to the decision boundary that naturally have low consistency in neighbours' predictions.
    % due to the closeness to the boundary.
    \cref{tab:linear_vs_exponential_positive_vs_negative} (Experiments Section), demonstrates the effectiveness of the exponential reweighting compared to a linear one.
    
    \subsection{Temporal Queue for Negative Pairs Exclusion}
    \label{subsubsec:temporal_queue_exclusion}
    As described in \cref{subsubsec:nnvoting}, the process of pseudo-labels refining through neighbours samples is based on the assumption that features vectors extracted from same-class samples lie closer in the features space rather than features from different-class samples. 
    To match this requirement, we use a contrastive framework on target data during the adaptation.
    Since target data is unlabelled, we employ a self-supervised contrastive training. Similar to \cite{moco}, we pull together features from different augmentations of the same image (positive pairs) and we push away features from other instances (negative pairs). For each sample $x_t$, we select two strong augmentations $t_{sa}, t'_{sa} \in \mathcal{T}_{sa}$ from the distribution $\mathcal{T}_{sa}$. Then, we generate two strongly-augmented samples $t_{sa}(x_t), t'_{sa}(x_t)$ and we encode them into query $q = f_t(t_{sa}(x_t))$ and key $k = f'_t(t'_{sa}(x_t))$ features. Queries and keys features will be used to build the positive pairs. To build negative pairs, we also maintain a queue $Q_e$ that stores keys features $\{ k^i\}_{i=1}^N$ computed in each mini-batch. The negative pairs are then composed by pairing features from the queue.
    
    \begin{figure}[t!]
    \includegraphics[width=\columnwidth]{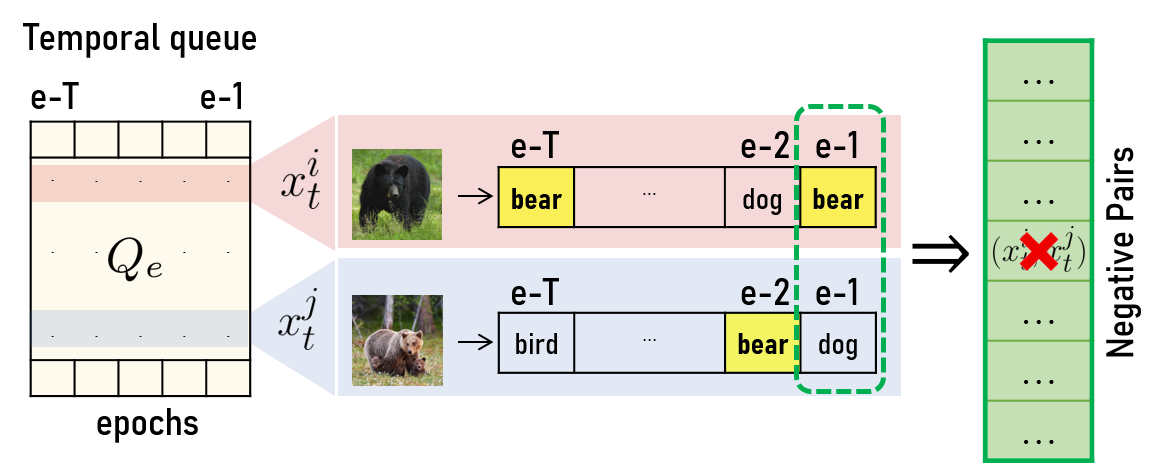}
        \vspace{-10pt}
    \centering
    \caption{A couple of target images would be wrongly considered as a negative pair if only comparing the latest predictions (green box). Instead, since $x_t^i$ and $x_t^j$ share the same pseudo-labels (at least once) in the past T epochs, i.e. $\{e-T,...,e-1 \}$, we exclude them from the list of negative pairs.} 
    \label{fig:queue_based_exclusion}
    \vspace{-10pt}
    \end{figure}
    
    MoCo \cite{moco} uses the pairs $(q,k)$ as positive and all the pairs $\{(q,k^i) \}_{i=1}^N$ as negative pairs by minimising and maximising their cosine distance. Since features are stored in $Q_e$ independently from the class, even features from samples sharing the same class will be pushed apart, which is in contrast with our objective.
    Recently, Chen et al. \cite{contrastive_testtime} propose a strategy to exclude some negative pairs from the contrastive loss. For every negative pair, they just compare the pseudo-labels of the two samples composing the pair. If the two samples share the same pseudo-label, then the negative pair is masked out. Otherwise, the negative pair composed of the two samples is included in the negative pairs list. 
    However, the exclusion strategy of \cite{contrastive_testtime} does not take into account the noise that inevitably affects the pseudo-labels during the adaptation. If one or both samples that compose the negative pair have noisy (incorrect) pseudo-labels, the comparison will be totally distorted. As a result, a high number of ``false negative'' pairs (pairs made by samples sharing the same class but with different pseudo-labels) will be wrongly included in the negative pairs list. 
    Hence, features from samples sharing the same class will be wrongly pushed away due to the noise.
    
    We introduce a novel negative pairs exclusion strategy able to identify and exclude pairs made of samples that belong to the same class, even in presence of noise in the pseudo-labels.
    Differently from \cite{contrastive_testtime}, which rely on current pseudo-labels only, the proposed exclusion strategy looks at the history of pseudo-labels that samples had during the training. By looking at the history of pseudo-labels, we have a higher probability to observe, at least one time, the correct label in the history rather than by looking only at the current pseudo-label. Hence, even if a sample has a noisy pseudo-label in the current epoch, the history will probably reveal its correct one and this allows us to correctly identify and exclude negative pairs made of samples sharing the same class. 
    
    To this end, we turn the queue $Q_e$ into a temporal queue by also storing for each key features the refined pseudo-labels $\{\hat{y}^j\}_{j=1}^T$ of the $T$ past epochs, i.e $\{e-T,...,e-1\}$.
    Then, we exclude from the negative pairs list all the pairs that shared the same pseudo-labels at least once in the past $T$ epochs, as illustrated in \cref{fig:queue_based_exclusion}.
    Accordingly, we optimise the following InfoNCELoss \cite{infonceloss}:
    \begin{equation}
    L_t^{ctr} = L_{\text{InfoNCE}} = - \, log \, \frac{exp(q \cdot k_+ / \tau)}{\sum_{j\in \mathcal{N}_q} exp(q \cdot k_j / \tau)}
    %\vspace{-7pt}
    \end{equation}
    $$
    \mathcal{N}_q = \{ j | \hat{y}_j^i \neq \hat{y}^i , \;\forall j \in \{1,...,N\}, \forall i\in\{1,...,T\} \}, %\vspace{-10pt}
    $$
    \noindent where $\mathcal{N}_q$ is the set of indices of samples in $Q_e$ that never shared with the query sample the same pseudo-labels in the past $T$ epochs.
    In \cref{subsec:analysis}, we analyse the behaviour of our method using different values for $T$. 
    
    \subsection{Joint training with self-learning}
    \label{subsec:overallframework}
    The refined pseudo-labels obtained with neighbours' knowledge aggregation are used to compute a classification loss on target data in order to adapt the model to the new domain.
    We use the refined pseudo-labels $\hat{y}_t$ obtained from a weakly-augmented image $t_x(x_t)$ as self-supervision for the strongly-augmented version $t_{sa}(x_t)$.
    
    The refining of the pseudo-labels is an iterative process, so it progressively improves the pseudo-labels accuracy during the training. This means that, mostly in the early stage of the training, some noise is still present in the pseudo-labels.
   
    In addition to the proposed reweighting and exclusion strategies, to mitigate the effect of the noisy pseudo-labels, we use the negative learning loss \cite{NLNL, JPNL} as classification loss.  Differently from \cite{NLNL, JPNL}, which use the negative learning loss concurrently or alternating with a standard positive loss, we do not use the positive loss in the entire training. In \cref{subsec:analysis}, we discuss the benefits given by using only the negative loss.
    As a result, our classification loss is the following:
    \begin{equation}
    \label{eq:cls_loss}
    L_t^{cls} = - \; \mathbb{E}_{x_t \in \mathcal{X}_t} \; \Bigr[ w_{x_t} \cdot \sum_{c=1}^C  \tilde{y}^c \, log \, (1-p_{sa}^c) \Bigr],
    \end{equation}
    where $\tilde{y}^c$ is a complementary label $\tilde{y} \in \{1,...,C\} \: \backslash \: \{\hat{y}_t\}$ chosen randomly from the set of labels and without the refined pseudo-labels, $p_{sa} = \sigma(g_t(t_{sa}(x_t)))$ is the probabilistic output for the strongly-augmented image $t_{sa}(x_t))$ and $w_{x_t}$ is the weight that estimates the uncertainty of $\hat{y}_t$, as explained in \cref{subsec:entropy_estimation}.
    The random selection of the complementary label $\tilde{y}$ is coherent with the negative learning framework \cite{NLNL}.
    
    To prevent the posterior collapse, we follow the standard state-of-the-art protocol by optimising the following regularisation term:
    $$
        L_t^{div} = \mathbb{E}_{x_t \in \mathcal{X}_t}  \sum_{c=1}^C  \bar{p}_q^c \, log \, \bar{p}_q^c,
        \quad
        \bar{p}_q = \mathbb{E}_{x_t \in \mathcal{X}_t} \; \sigma(g_t(t_{sa}(x_t))).
    $$
    
    \noindent The overall loss function used for the target data training is the following:
    $$
        L_t = \gamma_1 L_t^{cls} + \gamma_2 L_t^{ctr} + \gamma_3 L_t^{div},
    $$
    
    \noindent where $\gamma_1 = \gamma_2 = \gamma_3 = 1$ are non-tuned hyper-parameters.

\section{Experiments and Results}
\label{sec:experiments_results}

\subsection{Experimental setup}
\label{subsec:experimental_setup}
% \vspace{-10pt}
    \noindent\textbf{Datasets}. To evaluate the goodness of our approach, we employ experiments on PACS \cite{pacs}, VisDA-C \cite{visdac} and DomainNet \cite{domainnet}. \\
    \textit{\textbf{- PACS}} contains 4 domains (\textit{Art-Painting, Cartoon, Photo and Sketch}) and 7 object categories with a large domain shift due to different styles. We perform experiments for both single-source and multi-target settings and we compare the average results among domain combinations.\\
   \textit{ \textbf{- VisDA-C}} is a challenging large-scale dataset with a large synthetic-to-real domain gap across 12 object categories. We report and compare the per-class top-1 accuracy and their average (Avg.).\\
    \textit{\textbf{- DomainNet} }is a large-scale dataset. Following \cite{mini_domainnet}, we use a subset of it that contains 126 classes from 4 domains (\textit{Real, Sketch, Clipart, Painting}) and we refer to it as DomainNet-126. We evaluate 7 domain shifts built from the 4 domains and we report the top-1 accuracy under each domain shift as well as their average (Avg.).\\
    \textbf{Implementation details}.
    We use standard classification architectures comprising a feature extractor followed by a classifier. For fair comparison purposes, we choose the same ResNet 18/50/101 models \cite{resnet} used by competitors as backbones in the experiments. Specifically, we use ResNet18 for PACS, ResNet50 for DomainNet and ResNet101 for VisDA-C experiments (as detailed in the caption of each Table). Following SHOT \cite{sfda2}, we add an extra 256-dimensional fully-connected+BatchNorm bottleneck after the encoder output. For experiments on PACS and VisDA-C, we also apply WeightNorm \cite{weight_norm} on the classifier.

    For source training, we initialise the ResNet backbone with ImageNet-1K \cite{imagenet} pre-trained weights available in the Pytorch model zoo. We train the source model with the standard cross-entropy loss and with label-smoothing like in \cite{label_smoothing}. For the adaptation phase, the target model is initialised with the source model's parameters. 
    For more details, the code that is available at \href{https://github.com/MattiaLitrico/Guiding-Pseudo-labels-with-Uncertainty-Estimation-for-Source-free-Unsupervised-Domain-Adaptation}{https://github.com/MattiaLitrico/Guiding-Pseudo-labels-with-Uncertainty-Estimation-for-Source-free-Unsupervised-Domain-Adaptation}.

\begin{table}[t]
\begin{adjustbox}{width=\columnwidth}
\begin{tabular}{lc|ccc|ccc|g}
\toprule
       \multicolumn{9}{c}{Single-Source UDA} \\ \midrule
Method & SF & P $\rightarrow$ A  & P $\rightarrow$ C    & P $\rightarrow$ S & A $\rightarrow$ P   & A $\rightarrow$ C & A $\rightarrow$ S & Avg. 
\\ \midrule
NEL  \cite{nel}  & yes & 82.6 & 80.5 & 32.3 & 98.4 & 84.3 & 56.1 & 72.4                                      \\ \midrule
\textbf{Ours} & yes  & \textbf{87.5} & \textbf{84.2} & \textbf{75.8} & \textbf{98.8} & \textbf{84.6} & \textbf{77.2} & \textbf{84.7} \\ \bottomrule
\end{tabular}
\end{adjustbox}
\caption{Classification accuracy (\%) on PACS for the single-source setting. All methods use the ResNet-18 backbone. Highest accuracies are in bold. We surpass the NEL \cite{nel} baseline by 12.3\%.}
\label{tab:pacs_singlesource}
%\vspace{-10pt}
\end{table}

\begin{table}[t]
\begin{adjustbox}{width=\columnwidth}
\begin{tabular}{lc|ccc|ccc|g}
\toprule

% \multicolumn{9}{c}{Multi-Target UDA}  \\ \midrule
\multicolumn{2}{c|}{Multi-Target UDA}  & \multicolumn{3}{c|}{P $\rightarrow$ ACS }    & \multicolumn{3}{c|}{A $\rightarrow$ PCS }  \\ \midrule
Method & SF &  A  &  C    &  S & P   & C &  S & Avg. \\ \midrule
1-NN  & no &  15.2          & 18.1          & 25.6                   & 22.7          & 19.7          & 22.7        & 20.7                  \\
ADDA \cite{uda4} & no &  24.3          & 20.1          & 22.4                   & 32.5          & 17.6          & 18.9       & 22.6                  \\
DSN \cite{dsn}  & no &  28.4          & 21.1          & 25.6                   & 29.5          & 25.8          & 24.6         & 25.8                  \\
ITA  \cite{ita} & no & 31.4          & 23.0          & 28.2                   & 35.7          & 27.0          & 28.9         & 29.0                  \\ 
KD \cite{kd} & no & 24.6          & 32.2          & 33.8                   & 35.6          & 46.6          & 57.5         & 46.6                  \\\midrule
NEL   \cite{nel} & yes & \textbf{80.1} & \textbf{76.1}          & 25.9                   & 96.0          & \textbf{82.8}          & 49.8          & 68.4                  \\ \midrule
\textbf{Ours}   & yes & 74.7          & 70.1 & \textbf{68.7}          & \textbf{94.6} & 70.8 & \textbf{71.5} & \textbf{75.0}        \\ \bottomrule
\end{tabular}
\end{adjustbox}
\caption{Classification accuracy (\%) on PACS for the multi-target setting. All methods use the ResNet-18 backbone. Highest accuracies are in bold. We surpass the SF-UDA baseline NEL \cite{nel} by 6.6\%.}
\vspace{-1.5em}
\label{tab:pacs_multitarget}
\end{table}

\begin{table*}[t]
\tiny
\begin{adjustbox}{width=1\textwidth}
\begin{tabular}{lc|cccccccccccc|g}
\toprule
    
Method & SF-UDA & plane    & bcycl    & bus    & car   & horse    & knife    & mcycl    &
person    &
plant   &
sktbrd    &
train  &
truck   &
Avg.   
\\ [-1pt] \midrule

CDAN \cite{cdan} & no & 85.2 & 66.9 & 83.0 & 50.8 & 84.2 & 74.9 & 88.1 & 74.5
& 83.4 & 76.0 & 81.9 & 38.0 & 73.9
\\ 
CDAN+BSP \cite{cdan_bsp} & no & 92.4 & 61.0 & 81.0 & 57.5 & 89.0 & 80.6 & 90.1 & 77.0
& 84.2 & 77.9 & 82.1 & 38.4 & 75.9
\\ 
SWD \cite{swd} & no & 90.8 & 82.5 & 81.7 & 70.5 & 91.7 & 69.5 & 86.3 & 77.5
& 87.4 & 63.6 & 85.6 & 29.2 & 76.4
\\ 
MCC \cite{mcc} & no & 88.7 & 80.3 & 80.5 & 71.5 & 90.1 & 93.2 & 85.0 & 71.6
& 89.4 & 73.8 & 85.0 & 36.9 & 78.8 
\\
CAN \cite{uda6} & no & \textbf{97.0} & 87.2 & 82.5 & 74.3 & \textbf{97.8} & 96.2 & 90.8 & 80.7
& 96.6 & \textbf{96.3} & 87.5 & 59.9 & 87.2
\\ [-1pt] \midrule
DivideMix \cite{DivideMix} & yes & 95.0 & 82.4 & 85.3 & 78.1 & 94.2 & 90.3 & 90.1 & 81.3 & 92.5 & 91.9 & 91.2 & 60.8 & 86.1
\\ [-1pt] \midrule
MA \cite{generative} & yes & 94.8 & 73.4 & 68.8 & 74.8 & 93.1 & 95.4 & 88.6 & 84.7
& 89.1 & 84.7 & 83.5 & 48.1 & 81.6
\\ 
BAIT \cite{bait} & yes & 93.7 & 83.2 & 84.5 & 65.0 & 92.9 & 95.4 & 88.1 & 80.8
& 90.0 & 89.0 & 84.0 & 45.3 & 82.7
\\ 
SHOT \cite{sfda2} & yes & 95.3 & 87.5 & 78.7 & 55.6 & 94.1 & 94.2 & 81.4 & 80.0
& 91.8 & 90.7 & 86.5 & 59.8 & 83.0
\\ 

DIPE \cite{dipe} & yes & 95.2 & 87.6 & 78.8 & 55.9 & 93.9 & 95.0 & 84.1 & 81.7
& 92.1 & 88.9 & 85.4 & 58.0 & 83.1
\\ 
NEL \cite{nel} & yes & 94.5 & 60.8 & \textbf{92.3} & 87.3 & 87.3 & 93.2 & 87.6 & \textbf{91.1}
& 56.9 & 83.4 & 93.7 & \textbf{86.6} & 84.2 
\\ 
$A^2$ Net \cite{a2net} & yes & 94.0 & 87.8 & 85.6 & 66.8 & 93.7 & 95.1 & 85.8 & 81.2
& 91.6 & 88.2 & 86.5 & 56.0 & 84.3 
\\ 
G-SFDA \cite{sfda3}   & yes & 96.1 & 88.3 & 85.5 & 74.1 & 97.1 & 95.4 & 89.5 & 79.4
& 95.4 & 92.9 & 89.1 & 42.6 & 85.4 
\\ 
SFDA-DE \cite{sfda_de} & yes & 95.3 & \textbf{91.2} & 77.5 & 72.1 & 95.7 & \textbf{97.8} & 85.5 & 86.1
& 95.5 & 93.0 & 86.3 & 61.6 & 86.5 
\\ 
AdaContrast \cite{contrastive_testtime} & yes & \textbf{97.0} & 84.7 & 84.0 & 77.3 & 96.7 & 93.8 & \textbf{91.9}
& 84.8 & 94.3 & 93.1 & \textbf{94.1} & 49.7 & 86.8
\\ 
CoWA \cite{cowa} & yes & 96.8 & 90.3 & 87.0 & 67.4 & 97.2 & 96.6 & 90.4
& 87.3 & 95.6 & 95.5 & 91.8 & 62.5 & 88.2
\\  [-1pt] \midrule
\textbf{Ours} & yes & 97.3 & 96.2 & 90.5 & \textbf{91.8} & 90.0 & 94.2 & 87.4
& 87.7 & \textbf{97.0} & 84.3 & 93.0 & 81.0 & \textbf{90.0} 
\\ \bottomrule
\end{tabular}
\end{adjustbox}
\caption{Classification accuracy (\%) on VisDA-C synthetic $\rightarrow$ real. All methods use the ResNet-101 backbone. The proposed approach outperforms the UDA state-of-the-art by 2.8\% on average (Avg.) and the previous SF-UDA state-of-the-art by 1.8\% on average (Avg.)}
\label{tab:visdac}
\vspace{-1.5em}
\end{table*}

\begin{table}[t!]

\begin{adjustbox}{width=\columnwidth}
\scriptsize
\begin{tabular}{lc|ccccccc|g}
\toprule
    
Method & SF-UDA & R $\rightarrow$ C  & R $\rightarrow$ P    & P $\rightarrow$ C & C $\rightarrow$ S   & S $\rightarrow$ P & R $\rightarrow$ S    & P $\rightarrow$ R    &
Avg. 

\\ \midrule
MCC \cite{mcc} & no  & 44.8 & 65.7 & 41.9 & 34.9 & 47.3 & 35.3 & 72.4
& 48.9 
\\ \midrule
DivideMix \cite{DivideMix} & yes & 68.1 & 69.5 & 67.7 & 61.3 & 64.3 & 62.4 & 77.3
& 67.2
\\ \midrule
TENT \cite{sfda1} & yes  & 58.5 & 65.7 & 57.9 & 48.5 & 52.4 & 54.0 & 67.0
& 57.7 
\\ 
SHOT \cite{sfda2} & yes  & 67.7 & 68.4 & 66.9 & 60.1 & 66.1 & 59.9 & \textbf{80.8}
& 67.1
\\ 
AdaContrast \cite{contrastive_testtime} & yes  & 70.2 & 69.8 & 68.6 & 58.0 & 65.9 & 61.5 & 80.5
& 67.8 
\\ \midrule
\textbf{Ours} & yes & \textbf{74.2} & \textbf{70.4} & \textbf{68.8} & \textbf{64.0} & \textbf{67.5} & \textbf{65.7} & 76.5
& \textbf{69.6}  
\\ \bottomrule
\end{tabular}
\end{adjustbox}
\caption{Classification accuracy (\%) on 7 domain shifts of DomainNet-126. All methods use the ResNet-50 backbone. The proposed approach achieves the highest accuracy on 6 domain shifts and the highest accuracy on average (Avg.).}
\label{tab:domainnet}
\end{table}

\begin{table}[t!]
\begin{adjustbox}{width=\columnwidth}
\scriptsize
\begin{tabular}{cccccg}
\toprule
    
\shortstack{Pseudo-label \\ refinement} & \shortstack{Contrastive \\ regularisation} & \shortstack{Negative \\ learning} & \shortstack{Temporal-queue \\ exclusion} & \shortstack{Uncertainty \\ reweighting} & \shortstack{Avg. \\ Acc.}
\\ \midrule
\cmark & \xmark   &  \xmark  &  \xmark  &  \xmark  & 52.3
\\
\cmark & \cmark &  \xmark  &  \xmark  &  \xmark  & 78.9
\\
\cmark & \cmark & \cmark &  \xmark  &  \xmark  & 82.1
\\
\cmark & \cmark & \cmark & \cmark &  \xmark  & 85.8
\\
\cmark & \cmark & \cmark & \cmark & \cmark & 90.0

\\ \bottomrule
\end{tabular}
\end{adjustbox}
\caption{Ablation studies of sub-components of the proposed method measured by classification accuracy (\%) on VisDA-C. First row the pseudo-label refinement \cref{subsubsec:nnvoting}. Second row the contrastive regularisation \cref{subsubsec:temporal_queue_exclusion}. Third row the negative learning loss \cref{subsec:overallframework}. Fourth row the proposed temporal-queue exclusion \cref{subsubsec:temporal_queue_exclusion}. Fifth row the proposed uncertainty reweighting \cref{subsec:entropy_estimation}.}
\label{tab:ablation}
\vspace{-1.5em}
\end{table}

    \subsection{Results}
    \label{subsec:results}
% \vspace{-12pt}
    \noindent\textbf{PACS: single-source and multi-target SF-UDA}.
    In \cref{tab:pacs_singlesource} and \cref{tab:pacs_multitarget} we show results obtained on the PACS dataset for single-source and multi-target experiments, respectively. For the single-source setting, we evaluate our method over six standard domain shifts using one domain as source domain and one as target domain. We surpass the SOTA in every experiment and with a large margin of +12.3\% on average. For the multi-target setting, we built two target domains by merging either the art-painting, cartoon and sketch domains (ACP) or the photo, cartoon and sketch domains (PCS). Results show that our approach improves SOTA baseline by +6.6\% on average.\\
    \textbf{VisDA-C synthetic $\rightarrow$ real}.
    \cref{tab:visdac} compares our method with state-of-the-art unsupervised domain adaptation and Source-free adaptation methods on the VisDA-C dataset with the synthetic to real shift. For the UDA setting, our method overcomes the strong baseline CAN \cite{uda6} by +2.8\% and outperforms all the other baselines by an even larger margin, even though we do not use source data at all during Source-free adaptation. For the more challenging setting SF-UDA, we achieve the highest per-class average accuracy by a notable margin of +1.8\% on a recent baseline. In addition, our method achieves a comparable accuracy in every class, while all the baselines drastically fail for at least one class.\\
    \textbf{DomainNet-126}.
    In \cref{tab:domainnet} we show results on 7 domain shifts of the large-scale dataset DomainNet-126 comparing with both UDA and SF-UDA baselines. Even if our approach does not require accessing source data during the adaptation, we achieve an increment of performances of +20.7\% on the average performance when compared with the UDA method MCC \cite{mcc}. In addition, our method achieves the highest accuracy on 5 domain shifts and it outperforms state-of-the-art SF-UDA approaches by +1.8\% on average. We also obtain stable performance across domains, with accuracies always above 60.0\%, which makes our method more suitable in practical contexts thanks to its reliability in all settings.\\
    \textbf{Label noise removal baseline}. For both VisDA-C and DomainNet-126, we also compare our method with DivideMix \cite{DivideMix}, a strong label noise removal baseline. Results in \cref{tab:visdac} and \cref{tab:domainnet} show that despite DivideMix is competitive with some SF-UDA baselines, our approach improves performance by +3.9\% and +2.4\%. As also suggested in Ahmed et al. \cite{nel}, the noise in pseudo-labels (called shift-noise) caused by the domain shift is very skewed and thus difficult to handle by standard label noise algorithms.

    \subsection{Analysis}
        \label{subsec:analysis}
    \noindent\textbf{Ablation Study.} In \cref{tab:ablation}, we report the results of ablation studies for individual components of our pipeline on VisDA-C.
    First, we only exploit pseudo-labels refinement (\cref{subsubsec:nnvoting}) achieving a lower accuracy of 52.3\%. 

    By inserting the contrastive regularisation (second row of \cref{tab:ablation}), we boost the accuracy by +26.6\%. 
    The third row of the table shows the effectiveness of the negative learning loss (\cref{subsec:overallframework}) compared to a positive one, with an increment of performances of +3.2\%, while the fourth presents results obtained by enabling the temporal queue-based exclusion strategy introduced in \cref{subsubsec:temporal_queue_exclusion}, which brings another performance gain of +3.7\%. 
    Finally, in the last row, we show the boost in performance obtained by the proposed uncertainty reweighting approach (\cref{subsec:entropy_estimation}) which further boosts performance by 4.2\%. \\
   
\begin{figure}%
    \centering
%    {\includegraphics[width=0.45\columnwidth]{figures/Plot_number_of_neighbours.png} }
    {\includegraphics[width=0.9\columnwidth]{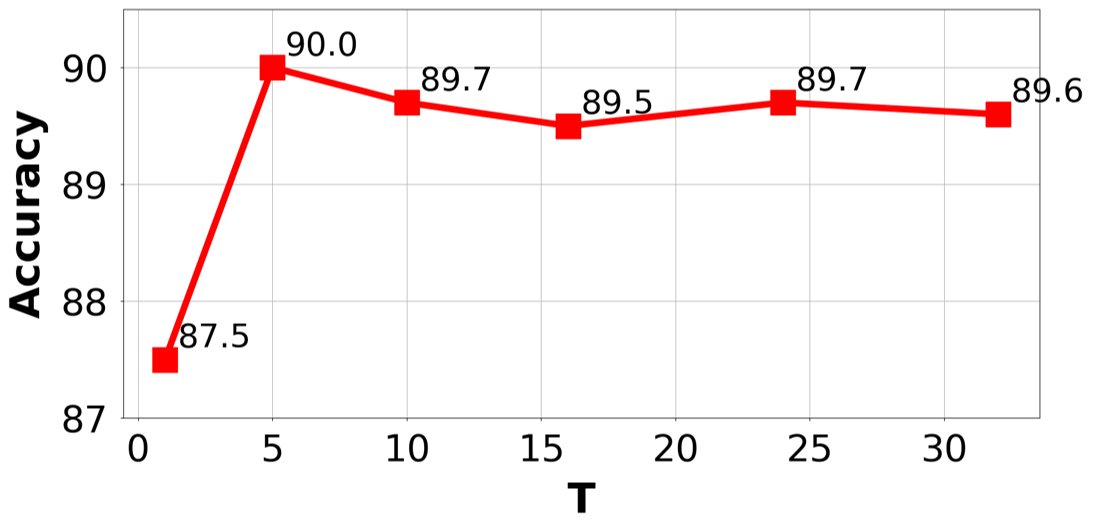} }

    \caption{Classification accuracy on VisDA-C (\%) versus length of the queue $Q_e$ in \cref{subsubsec:temporal_queue_exclusion}.}
    \label{fig:numneighbours_lengthqueue}%
    \end{figure}

\begin{figure}%
    \centering
    {\includegraphics[width=0.9\columnwidth]{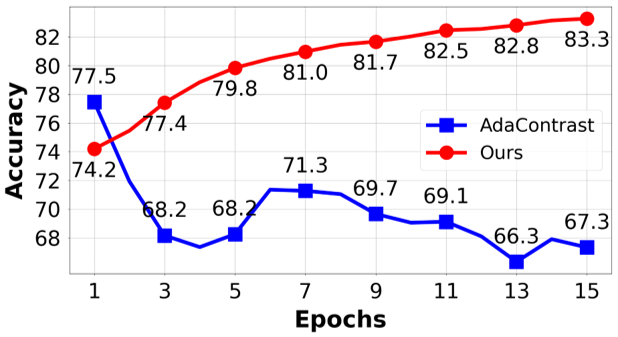} }
    \caption{Accuracy of the refined pseudo-labels on VisDA-C. The noise is not handled in AdaContrast \cite{contrastive_testtime} and the model progressively overfits wrong pseudo-labels. On the contrary, our reweighting and exclusion strategies mitigate the effects of noisy samples, resulting in progressively improving the pseudo-labels accuracy.}%
    \label{fig:acc_pseudolabels}%
    \end{figure}

    \noindent\textbf{Length of the Temporal Queue.}
    \cref{fig:numneighbours_lengthqueue} plots the trend of the accuracy with different values of $T$, i.e. the temporal length of the queue $Q_e$ (see \cref{subsubsec:temporal_queue_exclusion}). We achieve the best performance using $T=5$ that allows us to have a negligible memory overhead to memorise the $T$ past predictions. Note that the worst accuracy is obtained using $T=1$ that corresponds to a naive exclusion strategy using only current predictions.\\
    \textbf{Guiding the Pseudo-labels Refinement.}
    \cref{fig:acc_pseudolabels} plots the trend of the accuracy of the refined pseudo-labels during the adaptation. AdaContrast \cite{contrastive_testtime} is highly affected by the noise in the pseudo-labels resulting in progressively decreasing the pseudo-labels accuracy. On the contrary, \cref{fig:acc_pseudolabels} demonstrates the effectiveness of the proposed reweighting and exclusion strategies in increasing the robustness to the noise. As a consequence, our method is able to progressively improve the accuracy of pseudo-labels and thus guides the learning through more accurate labels.\\
    \textbf{Hard sample selection.} 
    We test a hard strategy based on entropy margin \cite{Iwasawa}: we set a threshold on the entropy and train the network only with samples above the threshold. Results in \cref{tab:linear_vs_exponential_positive_vs_negative} show that our smoother solution performs largely better. We hypothesize that the smoothness of our reweighting strategy allows to not highly penalise samples that inevitably have an high entropy, such as samples near the boundaries. \\
    \textbf{Linear vs Exponential Loss Reweighting.}
    \cref{tab:linear_vs_exponential_positive_vs_negative} (Top) demonstrates the effectiveness of using an exponential function in \cref{eq:reweighting} rather than a linear one. Although using a linear reweighting our method overcomes multiple baselines, with an accuracy of 85.1\%, the proposed exponential reweighting achieves a gain of performance of +4.9\%.\\
    \textbf{Negative vs Positive Learning.}
    \cref{tab:linear_vs_exponential_positive_vs_negative} (Bottom) shows results obtained optimising, as classification loss, only a standard positive loss (first row), a linear combination of a positive and a negative loss (second row), and only a negative learning loss. Although the noise affects the pseudo-labels, our method achieves satisfying performance even using a positive loss. This emphasises the role of our reweighting strategy in dealing with noise. Nonetheless, using only the negative loss we achieve a gain of performances of +4.8\%.
    \section{Limitations and Impact}
    \label{sec:limit}
    The proposed approach, if compared to other source-free UDA methods, comes with the reasonable overhead of maintaining and updating a queue during training. Concerning possible negative impact, our method shares potential downsides of any other UDA algorithm: while in general they can allow to reliably deploy AI systems in new domains, such systems could be subjected to inappropriate use or lead to harmful applications.
    
    \begin{table}[t!]
    \centering
    % \begin{adjustbox}{width=\columnwidth}
    % \scriptsize
    \renewcommand{\arraystretch}{0.2}
    %\resizebox{\columnwidth}{!}
    \begin{tabular}{p{5cm} c}
    \toprule
    Method & Acc. \\ \midrule

    Ours w/ hard entropy margin & 85.9
    \\ \midrule
    
    Ours w/ linear weighting & 85.1
    \\ 
    % \textbf{Ours w/ exponential weighting} & \textbf{90.0}
    \\ \midrule 
    Ours w/ positive & 83.0
    \\ 
    Ours w/ positive+negative & 85.2
    \\ \midrule \midrule
    \textbf{Ours} & \textbf{90.0}
    \\ \bottomrule
        \vspace{-10pt}
    \end{tabular}
    % \end{adjustbox}
    \caption{Classification accuracy (\%) on VisDA-C comparing linear vs exponential weighting in \cref{eq:reweighting} and positive vs negative classification loss.}
    \label{tab:linear_vs_exponential_positive_vs_negative}
    \vspace{-10pt}
    \end{table}

    \section{Conclusion}
    \label{sec:conclusion}
    In this work, we introduced a novel approach for SF-UDA in image classification. Our method aggregates knowledge from neighbours to refine pseudo-labels and estimates their uncertainty in order to mitigate the impact of wrong assignments. We also introduced a novel negative pairs exclusion strategy, used inside a contrastive framework, to identify and exclude pairs made of samples sharing the same class. Our method surpassed the SOTA by a large margin on major DA benchmarks. Additional analyses experimentally demonstrated the effectiveness of the introduced strategies in increasing the robustness against noise, as well as improving the accuracy of the pseudo-labels that results in guiding the adaptation with significantly more accurate labels. 

    \section*{Acknowledgement}
    \noindent We would like to thank all the PAVIS Lab for their fundamental contribution.
    A special thank to Davide Talon whose effort in reviewing this paper allowed us to highly increase the quality of the manuscript.

%%%%%%%%% REFERENCES
{\small
\bibliographystyle{ieee_fullname}
\bibliography{egbib}

\begin{thebibliography}{10}\itemsep=-1pt

\bibitem{nel}
Waqar Ahmed, Pietro Morerio, and Vittorio Murino.
\newblock Cleaning noisy labels by negative ensemble learning for source-free
  unsupervised domain adaptation.
\newblock In {\em 2022 IEEE/CVF Winter Conference on Applications of Computer
  Vision (WACV)}, pages 356--365, 2022.

\bibitem{uda1}
Konstantinos Bousmalis, Nathan Silberman, David Dohan, D. Erhan, and Dilip
  Krishnan.
\newblock Unsupervised pixel-level domain adaptation with generative
  adversarial networks.
\newblock {\em 2017 IEEE Conference on Computer Vision and Pattern Recognition
  (CVPR)}, pages 95--104, 2017.

\bibitem{dsn}
Konstantinos Bousmalis, George Trigeorgis, Nathan Silberman, Dilip Krishnan,
  and Dumitru Erhan.
\newblock Domain separation networks.
\newblock In {\em Proceedings of the 30th International Conference on Neural
  Information Processing Systems}, NIPS'16, page 343–351, Red Hook, NY, USA,
  2016. Curran Associates Inc.

\bibitem{self_supervised1}
Mathilde Caron, Piotr Bojanowski, Armand Joulin, and Matthijs Douze.
\newblock Deep clustering for unsupervised learning of visual features.
\newblock In Vittorio Ferrari, Martial Hebert, Cristian Sminchisescu, and Yair
  Weiss, editors, {\em Computer Vision -- ECCV 2018}, pages 139--156, Cham,
  2018. Springer International Publishing.

\bibitem{contrastive_testtime}
Dian Chen, Dequan Wang, Trevor Darrell, and Sayna Ebrahimi.
\newblock Contrastive test-time adaptation.
\newblock In {\em CVPR}, 2022.

\bibitem{cl1}
Ting Chen, Simon Kornblith, Mohammad Norouzi, and Geoffrey Hinton.
\newblock A simple framework for contrastive learning of visual
  representations.
\newblock In {\em Proceedings of the 37th International Conference on Machine
  Learning}, ICML'20. JMLR.org, 2020.

\bibitem{self_supervised2}
Ting Chen, Simon Kornblith, Kevin Swersky, Mohammad Norouzi, and Geoffrey
  Hinton.
\newblock Big self-supervised models are strong semi-supervised learners.
\newblock In {\em Proceedings of the 34th International Conference on Neural
  Information Processing Systems}, NIPS'20, Red Hook, NY, USA, 2020. Curran
  Associates Inc.

\bibitem{cl2}
Xinlei Chen and Kaiming He.
\newblock Exploring simple siamese representation learning.
\newblock In {\em 2021 IEEE/CVF Conference on Computer Vision and Pattern
  Recognition (CVPR)}, pages 15745--15753, 2021.

\bibitem{cdan_bsp}
Xinyang Chen, Sinan Wang, Mingsheng Long, and Jianmin Wang.
\newblock Transferability vs. discriminability: Batch spectral penalization for
  adversarial domain adaptation.
\newblock In Kamalika Chaudhuri and Ruslan Salakhutdinov, editors, {\em
  Proceedings of the 36th International Conference on Machine Learning},
  volume~97 of {\em Proceedings of Machine Learning Research}, pages
  1081--1090. PMLR, 09--15 Jun 2019.

\bibitem{imagenet}
Jia Deng, Wei Dong, Richard Socher, Li-Jia Li, Kai Li, and Li Fei-Fei.
\newblock Imagenet: A large-scale hierarchical image database.
\newblock In {\em 2009 IEEE Conference on Computer Vision and Pattern
  Recognition}, pages 248--255, 2009.

\bibitem{sfda_de}
Ning Ding, Yixing Xu, Yehui Tang, Chao Xu, Yunhe Wang, and Dacheng Tao.
\newblock Source-free domain adaptation via distribution estimation.
\newblock In {\em 2022 IEEE/CVF Conference on Computer Vision and Pattern
  Recognition (CVPR)}, pages 7202--7212, 2022.

\bibitem{adversarial}
Yaroslav Ganin and Victor Lempitsky.
\newblock Unsupervised domain adaptation by backpropagation.
\newblock In {\em Proceedings of the 32nd International Conference on
  International Conference on Machine Learning - Volume 37}, ICML'15, page
  1180–1189. JMLR.org, 2015.

\bibitem{uda2}
Yaroslav Ganin, Evgeniya Ustinova, Hana Ajakan, Pascal Germain, Hugo
  Larochelle, Fran\c{c}ois Laviolette, Mario Marchand, and Victor Lempitsky.
\newblock Domain-adversarial training of neural networks.
\newblock {\em J. Mach. Learn. Res.}, 17(1):2096–2030, jan 2016.

\bibitem{ita}
Behnam Gholami, Pritish Sahu, Ognjen Rudovic, Konstantinos Bousmalis, and
  Vladimir Pavlovic.
\newblock Unsupervised multi-target domain adaptation: An information theoretic
  approach.
\newblock {\em IEEE Transactions on Image Processing}, 29:3993--4002, 2020.

\bibitem{MAEGhosh}
Aritra Ghosh, Himanshu Kumar, and P.~S. Sastry.
\newblock Robust loss functions under label noise for deep neural networks.
\newblock In {\em Proceedings of the Thirty-First AAAI Conference on Artificial
  Intelligence}, AAAI'17, page 1919–1925. AAAI Press, 2017.

\bibitem{self_supervised3}
Spyros Gidaris, Praveer Singh, and Nikos Komodakis.
\newblock Unsupervised representation learning by predicting image rotations.
\newblock {\em ArXiv}, abs/1803.07728, 2018.

\bibitem{noise_transition_matrix}
Jacob Goldberger and Ehud Ben-Reuven.
\newblock Training deep neural-networks using a noise adaptation layer.
\newblock In {\em ICLR}, 2017.

\bibitem{domain_shift2}
A. Gretton, AJ. Smola, J. Huang, M. Schmittfull, KM. Borgwardt, and B.
  Sch{\"o}lkopf.
\newblock {\em Covariate shift and local learning by distribution matching},
  pages 131--160.
\newblock MIT Press, Cambridge, MA, USA, 2009.

\bibitem{cl3}
Jean-Bastien Grill, Florian Strub, Florent Altch\'{e}, Corentin Tallec,
  Pierre~H. Richemond, Elena Buchatskaya, Carl Doersch, Bernardo~Avila Pires,
  Zhaohan~Daniel Guo, Mohammad~Gheshlaghi Azar, Bilal Piot, Koray Kavukcuoglu,
  R\'{e}mi Munos, and Michal Valko.
\newblock Bootstrap your own latent a new approach to self-supervised learning.
\newblock In {\em Proceedings of the 34th International Conference on Neural
  Information Processing Systems}, NIPS'20, Red Hook, NY, USA, 2020. Curran
  Associates Inc.

\bibitem{moco}
Kaiming He, Haoqi Fan, Yuxin Wu, Saining Xie, and Ross Girshick.
\newblock Momentum contrast for unsupervised visual representation learning.
\newblock In {\em 2020 IEEE/CVF Conference on Computer Vision and Pattern
  Recognition (CVPR)}, pages 9726--9735, 2020.

\bibitem{resnet}
Kaiming He, Xiangyu Zhang, Shaoqing Ren, and Jian Sun.
\newblock Deep residual learning for image recognition.
\newblock In {\em 2016 IEEE Conference on Computer Vision and Pattern
  Recognition (CVPR)}, pages 770--778, 2016.

\bibitem{uda3}
Judy Hoffman, Eric Tzeng, Taesung Park, Jun-Yan Zhu, Phillip Isola, Kate
  Saenko, Alexei Efros, and Trevor Darrell.
\newblock {C}y{CADA}: Cycle-consistent adversarial domain adaptation.
\newblock In Jennifer Dy and Andreas Krause, editors, {\em Proceedings of the
  35th International Conference on Machine Learning}, volume~80 of {\em
  Proceedings of Machine Learning Research}, pages 1989--1998. PMLR, 10--15 Jul
  2018.

\bibitem{Iwasawa}
Yusuke Iwasawa and Yutaka Matsuo.
\newblock Test-time classifier adjustment module for model-agnostic domain
  generalization.
\newblock In M. Ranzato, A. Beygelzimer, Y. Dauphin, P.S. Liang, and J.~Wortman
  Vaughan, editors, {\em Advances in Neural Information Processing Systems},
  volume~34, pages 2427--2440. Curran Associates, Inc., 2021.

\bibitem{reweighting3}
Lu Jiang, Zhenyuan Zhou, Thomas Leung, Jia Li, and Fei-Fei Li.
\newblock Mentornet: Learning data-driven curriculum for very deep neural
  networks on corrupted labels.
\newblock In {\em ICML}, 2018.

\bibitem{mcc}
Ying Jin, Ximei Wang, Mingsheng Long, and Jianmin Wang.
\newblock Minimum class confusion for versatile domain adaptation.
\newblock In Andrea Vedaldi, Horst Bischof, Thomas Brox, and Jan-Michael Frahm,
  editors, {\em Computer Vision -- ECCV 2020}, pages 464--480, Cham, 2020.
  Springer International Publishing.

\bibitem{uda6}
Guoliang Kang, Lu Jiang, Yi Yang, and Alexander~G. Hauptmann.
\newblock Contrastive adaptation network for unsupervised domain adaptation.
\newblock In {\em 2019 IEEE/CVF Conference on Computer Vision and Pattern
  Recognition (CVPR)}, pages 4888--4897, 2019.

\bibitem{NLNL}
Youngdong Kim, Junho Yim, Juseung Yun, and Junmo Kim.
\newblock Nlnl: Negative learning for noisy labels.
\newblock {\em 2019 IEEE/CVF International Conference on Computer Vision
  (ICCV)}, pages 101--110, 2019.

\bibitem{JPNL}
Youngdong Kim, Juseung Yun, Hyounguk Shon, and Junmo Kim.
\newblock Joint negative and positive learning for noisy labels.
\newblock In {\em Proceedings of the IEEE/CVF Conference on Computer Vision and
  Pattern Recognition (CVPR)}, pages 9442--9451, June 2021.

\bibitem{self_supervised4}
Gustav Larsson, Michael Maire, and Gregory Shakhnarovich.
\newblock Colorization as a proxy task for visual understanding.
\newblock {\em 2017 IEEE Conference on Computer Vision and Pattern Recognition
  (CVPR)}, pages 840--849, 2017.

\bibitem{swd}
Chen-Yu Lee, Tanmay Batra, Mohammad~Haris Baig, and Daniel Ulbricht.
\newblock Sliced wasserstein discrepancy for unsupervised domain adaptation.
\newblock In {\em 2019 IEEE/CVF Conference on Computer Vision and Pattern
  Recognition (CVPR)}, pages 10277--10287, 2019.

\bibitem{pseudolabel_1}
Dong-Hyun Lee.
\newblock Pseudo-label : The simple and efficient semi-supervised learning
  method for deep neural networks.
\newblock 2013.

\bibitem{cowa}
Jonghyun Lee, Dahuin Jung, Junho Yim, and Sungroh Yoon.
\newblock Confidence score for source-free unsupervised domain adaptation.
\newblock In Kamalika Chaudhuri, Stefanie Jegelka, Le Song, Csaba Szepesvari,
  Gang Niu, and Sivan Sabato, editors, {\em Proceedings of the 39th
  International Conference on Machine Learning}, volume 162 of {\em Proceedings
  of Machine Learning Research}, pages 12365--12377. PMLR, 17--23 Jul 2022.

\bibitem{reweighting4}
Kuang-Huei Lee, Xiaodong He, Lei Zhang, and Linjun Yang.
\newblock Cleannet: Transfer learning for scalable image classifier training
  with label noise.
\newblock {\em 2018 IEEE/CVF Conference on Computer Vision and Pattern
  Recognition}, pages 5447--5456, 2018.

\bibitem{pacs}
Da Li, Yongxin Yang, Yi-Zhe Song, and Timothy~M. Hospedales.
\newblock Deeper, broader and artier domain generalization.
\newblock {\em 2017 IEEE International Conference on Computer Vision (ICCV)},
  pages 5543--5551, 2017.

\bibitem{DivideMix}
Junnan Li, Richard Socher, and Steven C.~H. Hoi.
\newblock Dividemix: Learning with noisy labels as semi-supervised learning.
\newblock {\em ArXiv}, abs/2002.07394, 2020.

\bibitem{generative}
Rui Li, Qianfen Jiao, Wenming Cao, Hau-San Wong, and Si Wu.
\newblock Model adaptation: Unsupervised domain adaptation without source data.
\newblock In {\em 2020 IEEE/CVF Conference on Computer Vision and Pattern
  Recognition (CVPR)}, pages 9638--9647, 2020.

\bibitem{bait}
Rui Li, Qianfen Jiao, Wenming Cao, Hau-San Wong, and Si Wu.
\newblock Model adaptation: Unsupervised domain adaptation without source data.
\newblock In {\em Proceedings of the IEEE/CVF Conference on Computer Vision and
  Pattern Recognition (CVPR)}, June 2020.

\bibitem{sfda2}
Jian Liang, Dapeng Hu, and Jiashi Feng.
\newblock Do we really need to access the source data? source hypothesis
  transfer for unsupervised domain adaptation.
\newblock In {\em Proceedings of the 37th International Conference on Machine
  Learning}, ICML'20. JMLR.org, 2020.

\bibitem{pseudolabel_2}
Jian Liang, Dapeng Hu, and Jiashi Feng.
\newblock Domain adaptation with auxiliary target domain-oriented classifier.
\newblock In {\em 2021 IEEE/CVF Conference on Computer Vision and Pattern
  Recognition (CVPR)}, pages 16627--16637, 2021.

\bibitem{reweighting}
Tongliang Liu and Dacheng Tao.
\newblock Classification with noisy labels by importance reweighting.
\newblock {\em IEEE Transactions on Pattern Analysis and Machine Intelligence},
  38(3):447--461, 2016.

\bibitem{mmd1}
Mingsheng Long, Yue Cao, Jianmin Wang, and Michael~I. Jordan.
\newblock Learning transferable features with deep adaptation networks.
\newblock In {\em Proceedings of the 32nd International Conference on
  International Conference on Machine Learning - Volume 37}, ICML'15, page
  97–105. JMLR.org, 2015.

\bibitem{cdan}
Mingsheng Long, ZHANGJIE CAO, Jianmin Wang, and Michael~I Jordan.
\newblock Conditional adversarial domain adaptation.
\newblock In S. Bengio, H. Wallach, H. Larochelle, K. Grauman, N. Cesa-Bianchi,
  and R. Garnett, editors, {\em Advances in Neural Information Processing
  Systems}, volume~31. Curran Associates, Inc., 2018.

\bibitem{NCE}
Xingjun Ma, Hanxun Huang, Yisen Wang, Simone Romano, Sarah Erfani, and James
  Bailey.
\newblock Normalized loss functions for deep learning with noisy labels.
\newblock In Hal~Daumé III and Aarti Singh, editors, {\em Proceedings of the
  37th International Conference on Machine Learning}, volume 119 of {\em
  Proceedings of Machine Learning Research}, pages 6543--6553. PMLR, 13--18 Jul
  2020.

\bibitem{softvoting}
Harvey~B. Mitchell and Paul~A. Schaefer.
\newblock A “soft” k‐nearest neighbor voting scheme.
\newblock {\em International Journal of Intelligent Systems}, 16, 2001.

\bibitem{generative2}
Jogendra Nath~Kundu, Naveen Venkat, M.~V. Rahul, and R. Venkatesh~Babu.
\newblock Universal source-free domain adaptation.
\newblock In {\em 2020 IEEE/CVF Conference on Computer Vision and Pattern
  Recognition (CVPR)}, pages 4543--4552, 2020.

\bibitem{kd}
Le~Thanh Nguyen-Meidine, Atif Belal, M. Kiran, Jos{\'e} Dolz, Louis-Antoine
  Blais-Morin, and {\'E}ric Granger.
\newblock Knowledge distillation methods for efficient unsupervised adaptation
  across multiple domains.
\newblock {\em Image Vis. Comput.}, 108:104096, 2021.

\bibitem{self_supervised5}
Mehdi Noroozi and Paolo Favaro.
\newblock Unsupervised learning of visual representations by solving jigsaw
  puzzles.
\newblock In {\em ECCV}, 2016.

\bibitem{domainnet}
Xingchao Peng, Qinxun Bai, Xide Xia, Zijun Huang, Kate Saenko, and Bo Wang.
\newblock Moment matching for multi-source domain adaptation.
\newblock In {\em Proceedings of the IEEE International Conference on Computer
  Vision}, pages 1406--1415, 2019.

\bibitem{visdac}
Xingchao Peng, Ben Usman, Neela Kaushik, Judy Hoffman, Dequan Wang, and Kate
  Saenko.
\newblock Visda: The visual domain adaptation challenge.
\newblock {\em CoRR}, abs/1710.06924, 2017.

\bibitem{domain_shift}
Joaquin Quionero-Candela, Masashi Sugiyama, Anton Schwaighofer, and Neil~D.
  Lawrence.
\newblock {\em Dataset Shift in Machine Learning}.
\newblock The MIT Press, 2009.

\bibitem{reweighting2}
Mengye Ren, Wenyuan Zeng, Binh Yang, and Raquel Urtasun.
\newblock Learning to reweight examples for robust deep learning.
\newblock In {\em ICML}, 2018.

\bibitem{mini_domainnet}
Kuniaki Saito, Donghyun Kim, Stan Sclaroff, Trevor Darrell, and Kate Saenko.
\newblock Semi-supervised domain adaptation via minimax entropy.
\newblock {\em 2019 IEEE/CVF International Conference on Computer Vision
  (ICCV)}, pages 8049--8057, 2019.

\bibitem{self_supervised6}
Kuniaki Saito, Donghyun Kim, Stan Sclaroff, and Kate Saenko.
\newblock Universal domain adaptation through self-supervision.
\newblock In {\em Proceedings of the 34th International Conference on Neural
  Information Processing Systems}, NIPS'20, Red Hook, NY, USA, 2020. Curran
  Associates Inc.

\bibitem{self_supervised7}
Kuniaki Saito, Donghyun Kim, Stan Sclaroff, and Kate Saenko.
\newblock Universal domain adaptation through self supervision.
\newblock In H. Larochelle, M. Ranzato, R. Hadsell, M.F. Balcan, and H. Lin,
  editors, {\em Advances in Neural Information Processing Systems}, volume~33,
  pages 16282--16292. Curran Associates, Inc., 2020.

\bibitem{weight_norm}
Tim Salimans and Durk~P Kingma.
\newblock Weight normalization: A simple reparameterization to accelerate
  training of deep neural networks.
\newblock In D. Lee, M. Sugiyama, U. Luxburg, I. Guyon, and R. Garnett,
  editors, {\em Advances in Neural Information Processing Systems}, volume~29.
  Curran Associates, Inc., 2016.

\bibitem{fixmatch}
Kihyuk Sohn, David Berthelot, Chun-Liang Li, Zizhao Zhang, Nicholas Carlini,
  Ekin~D. Cubuk, Alex Kurakin, Han Zhang, and Colin Raffel.
\newblock Fixmatch: Simplifying semi-supervised learning with consistency and
  confidence.
\newblock In {\em Proceedings of the 34th International Conference on Neural
  Information Processing Systems}, NIPS'20, Red Hook, NY, USA, 2020. Curran
  Associates Inc.

\bibitem{sfda}
Yu Sun, Xiaolong Wang, Zhuang Liu, John Miller, Alexei Efros, and Moritz Hardt.
\newblock Test-time training with self-supervision for generalization under
  distribution shifts.
\newblock In Hal~Daumé III and Aarti Singh, editors, {\em Proceedings of the
  37th International Conference on Machine Learning}, volume 119 of {\em
  Proceedings of Machine Learning Research}, pages 9229--9248. PMLR, 13--18 Jul
  2020.

\bibitem{label_smoothing}
Christian Szegedy, Vincent Vanhoucke, Sergey Ioffe, Jon Shlens, and Zbigniew
  Wojna.
\newblock Rethinking the inception architecture for computer vision.
\newblock In {\em 2016 IEEE Conference on Computer Vision and Pattern
  Recognition (CVPR)}, pages 2818--2826, 2016.

\bibitem{uda4}
Eric Tzeng, Judy Hoffman, Kate Saenko, and Trevor Darrell.
\newblock Adversarial discriminative domain adaptation.
\newblock In {\em 2017 IEEE Conference on Computer Vision and Pattern
  Recognition (CVPR)}, pages 2962--2971, 2017.

\bibitem{mmd2}
Eric Tzeng, Judy Hoffman, Ning Zhang, Kate Saenko, and Trevor Darrell.
\newblock Deep domain confusion: Maximizing for domain invariance, 2014.

\bibitem{infonceloss}
A{\"a}ron van~den Oord, Yazhe Li, and Oriol Vinyals.
\newblock Representation learning with contrastive predictive coding.
\newblock {\em ArXiv}, abs/1807.03748, 2018.

\bibitem{ontarget}
Dequan Wang, Shaoteng Liu, Sayna Ebrahimi, Evan Shelhamer, and Trevor Darrell.
\newblock On-target adaptation, 2022.

\bibitem{sfda1}
Dequan Wang, Evan Shelhamer, Shaoteng Liu, Bruno Olshausen, and Trevor Darrell.
\newblock Tent: Fully test-time adaptation by entropy minimization.
\newblock In {\em International Conference on Learning Representations}, 2021.

\bibitem{dipe}
Fan Wang, Zhongyi Han, Yongshun Gong, and Yilong Yin.
\newblock Exploring domain-invariant parameters for source free domain
  adaptation.
\newblock In {\em 2022 IEEE/CVF Conference on Computer Vision and Pattern
  Recognition (CVPR)}, pages 7141--7150, 2022.

\bibitem{noise_robust_loss}
Yisen Wang, Xingjun Ma, Zaiyi Chen, Yuan Luo, Jinfeng Yi, and James Bailey.
\newblock Symmetric cross entropy for robust learning with noisy labels.
\newblock {\em 2019 IEEE/CVF International Conference on Computer Vision
  (ICCV)}, pages 322--330, 2019.

\bibitem{sample_selection}
Hongxin Wei, Lei Feng, Xiangyu Chen, and Bo An.
\newblock Combating noisy labels by agreement: A joint training method with
  co-regularization.
\newblock In {\em 2020 IEEE/CVF Conference on Computer Vision and Pattern
  Recognition (CVPR)}, pages 13723--13732, 2020.

\bibitem{a2net}
Haifeng Xia, Handong Zhao, and Zhengming Ding.
\newblock Adaptive adversarial network for source-free domain adaptation.
\newblock In {\em 2021 IEEE/CVF International Conference on Computer Vision
  (ICCV)}, pages 8990--8999, 2021.

\bibitem{sfda3}
Shiqi Yang, Yaxing Wang, Joost van~de Weijer, Luis Herranz, and Shangling Jui.
\newblock Generalized source-free domain adaptation.
\newblock In {\em 2021 IEEE/CVF International Conference on Computer Vision
  (ICCV)}, pages 8958--8967, 2021.

\bibitem{noise_overfitting}
Chiyuan Zhang, Samy Bengio, Moritz Hardt, Benjamin Recht, and Oriol Vinyals.
\newblock Understanding deep learning (still) requires rethinking
  generalization.
\newblock {\em Commun. ACM}, 64(3):107–115, feb 2021.

\end{thebibliography}
}

\end{document}